\documentclass[12pt]{article}
\usepackage{graphicx}
\usepackage{amsmath}
\usepackage{amssymb}
\usepackage{physics}
\usepackage{authblk}
\usepackage{amsthm}
\usepackage{booktabs}
\usepackage{algorithm}
\usepackage{algpseudocode}

\usepackage{geometry}
\usepackage{setspace}
\usepackage{titlesec}
\usepackage[numbers]{natbib}  
\usepackage[colorlinks=true, linkcolor=black, citecolor=blue]{hyperref}

\usepackage{hyperref}   
\usepackage{cleveref}
\crefname{figure}{Fig.}{Figs.}
\Crefname{figure}{Fig.}{Figs.}

\titleformat{\section}[block]{\normalfont\Large\bfseries}{\thesection}{1em}{}
\titleformat{\subsection}[block]{\normalfont\large\bfseries}{\thesubsection}{1em}{}
\titleformat{\subsubsection}[block]{\normalfont\normalsize\bfseries}{\thesubsubsection}{1em}{}

\title{\textbf{A Benchmark Suite for Multi-Objective Optimization in Battery Thermal Management System Design}}

\author[1]{\small Kaichen Ouyang\thanks{Correspondence: \texttt{oykc@mail.ustc.edu.cn}}}
\author[1]{\small Yezhi Xia\thanks{Correspondence: \texttt{xyz27@mail.ustc.edu.cn}}}
\affil[1]{\small University of Science and Technology of China, Hefei 230026, China}
\date{}

\begin{document}
\maketitle
\vspace{-1em}  

\begin{abstract}
Synthetic Benchmark Problems (SBPs) are commonly used to evaluate the performance of metaheuristic algorithms. However, these SBPs often contain various unrealistic properties, potentially leading to underestimation or overestimation of algorithmic performance. While several benchmark suites comprising real-world problems have been proposed for various types of metaheuristics, a notable gap exists for Constrained Multi-objective Optimization Problems (CMOPs) derived from practical engineering applications, particularly in the domain of Battery Thermal Management System (BTMS) design. To address this gap, this study develops and presents a specialized benchmark suite for multi-objective optimization in BTMS. This suite comprises a diverse collection of real-world constrained problems, each defined via accurate surrogate models based on recent research to efficiently represent complex thermal-fluid interactions. The primary goal of this benchmark suite is to provide a practical and relevant testing ground for evolutionary algorithms and optimization methods focused on energy storage thermal management. Future work will involve establishing comprehensive baseline results using state-of-the-art algorithms, conducting comparative analyses, and developing a standardized ranking scheme to facilitate robust performance assessment.
\end{abstract}


\section{Introduction}

Over the past decades, Constrained Multi-Objective Optimization Problems (CMOPs) have garnered significant attention, as the majority of optimization challenges in real-world applications are inherently subject to various constraints\cite{kumar2021benchmark,wei2025multiple,ouyang2025multi}. Typically, a CMOP involves simultaneously optimizing multiple conflicting objectives while satisfying a set of constraints, posing a considerable challenge for Evolutionary Algorithms (EAs) to find satisfactory trade-offs.

The performance evaluation of metaheuristic algorithms, given their stochastic nature, largely relies on empirical assessment using benchmark problems. Synthetic Benchmark Problems (SBPs) have been the conventional choice for this purpose, with several test suites being widely adopted\cite{tian2017platemo}. These artificial problems offer advantages such as simple mathematical formulation, low computational cost, known Pareto fronts, and scalability in objectives, variables, and constraints.

However, SBPs often suffer from a critical drawback: they may contain synthetic properties and regularities rarely encountered in practical engineering scenarios. This can lead to overestimation or underestimation of an algorithm's performance, potentially misleading the development and evaluation of Constrained Multi-Objective Evolutionary Algorithms (CMOEAs). For instance, the prevalence of problems with regular Pareto fronts in many suites might favor certain decomposition-based algorithms, failing to reflect the performance in more complex, real-world situations.

This gap underscores the necessity for benchmark suites composed of real-world problems. While such suites have been proposed for various classes of optimization algorithms, a notable absence exists for Constrained Multi-Objective Optimization (CMOO), particularly within the critically important domain of Battery Thermal Management System (BTMS) design\cite{zhou2024machine}. The effective thermal management of lithium-ion batteries is paramount for safety, performance, and longevity, and its design naturally presents complex, constrained, multi-objective optimization challenges involving conflicting goals like minimizing maximum temperature, temperature difference, pressure drop, and system weight.

To bridge this gap, this paper introduces a novel benchmark suite specifically tailored for multi-objective optimization in BTMS design. The primary contribution of this work is the development and presentation of a curated collection of Real-World Constrained Multi-Objective Optimization Problems (RWCMOPs) derived from recent and diverse BTMS research. A key feature of this suite is that each problem is defined via an accurate surrogate model—a mathematically simple yet effective approximation of complex, computationally expensive Computational Fluid Dynamics (CFD) simulations or detailed physical models. These surrogate models, often formulated using Response Surface Methodology (RSM) or other fitting techniques based on underlying experimental or high-fidelity numerical data, capture the essential thermal-fluid interactions and system behaviors, making the problems easy to implement and computationally inexpensive to evaluate.

This benchmark suite aims to provide a practical, relevant, and accessible testing ground for researchers to assess and compare the performance of CMOEAs on problems that reflect the true complexities and challenges of a vital engineering field. By offering a diverse set of problems with varying difficulty levels and characteristics, all defined within this document, we hope to facilitate more robust, reliable, and realistic performance assessments, thereby fostering advancements in optimization algorithms applicable to energy storage thermal management.

The remainder of this paper is organized as follows. Section 2 details the proposed BTMS benchmark suite, presenting the mathematical formulations of the RWCMOPs based on surrogate models. Finally, Section 3 concludes the paper and discusses potential future work.
\section{Battery Thermal Management System Design Test-Suite}
This section presents a comprehensive collection of real-world constrained multi-objective optimization problems (RWCMOPs) specifically curated for battery thermal management system (BTMS) design. The benchmark suite comprises 12 distinct BTMS configurations, each formulated as a multi-objective optimization problem with explicit mathematical expressions derived from surrogate models. These problems capture the essential trade-offs in BTMS design, including thermal performance (maximum temperature and temperature uniformity), hydraulic performance (pressure drop), and structural characteristics (system weight and size). All problems are presented in a standardized format with clearly defined decision variables, objective functions, and constraint boundaries, providing researchers with readily implementable test problems that reflect the complexities of practical BTMS design challenges.
\subsection{Serpentine Channel Liquid Cooling Optimization Model \protect\cite{en15239180}}
\begin{figure}[htbp]
    \centering
    \includegraphics[width=0.7\linewidth]{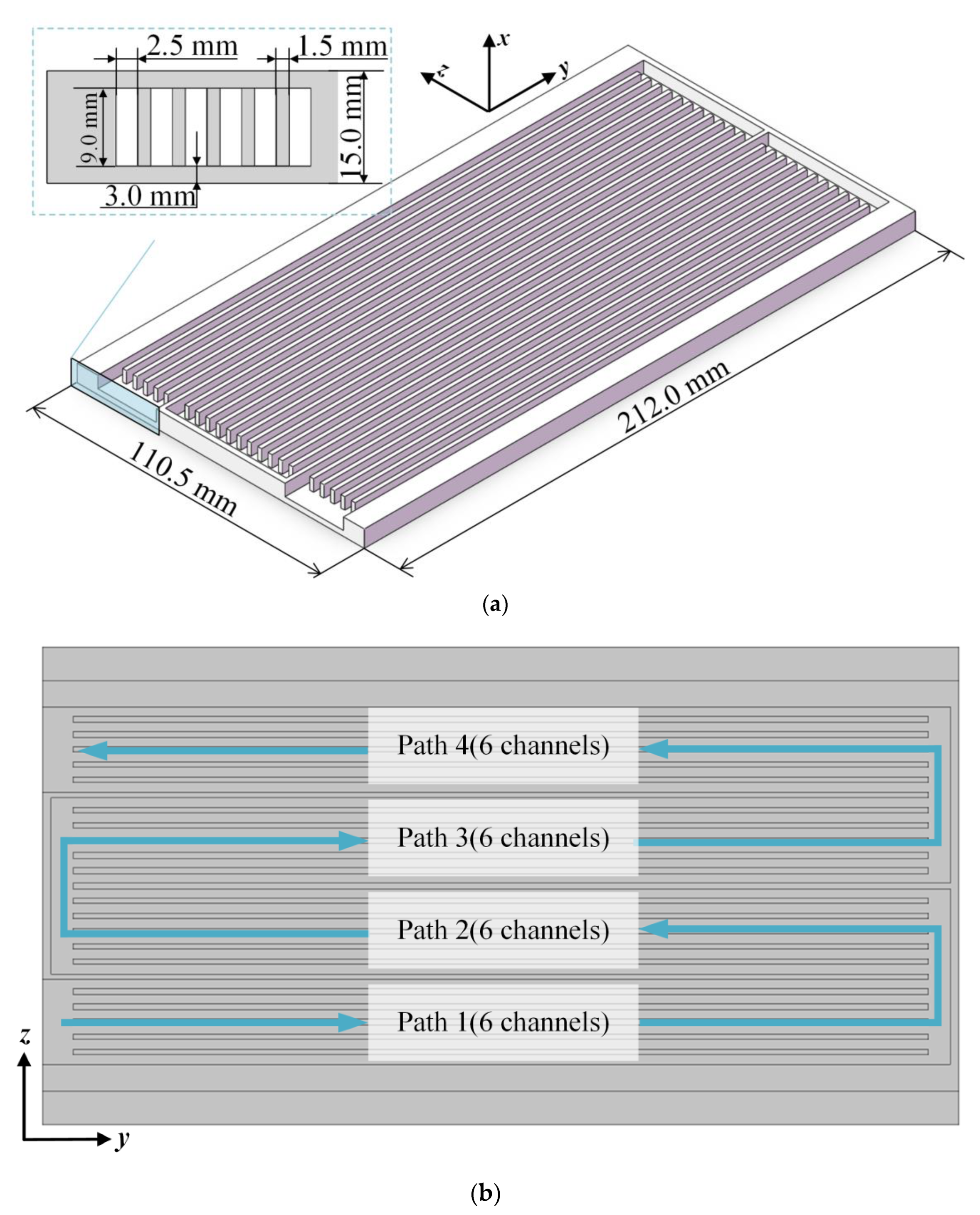}
    \caption{Schematic diagram of liquid cold plates with serpentine flow paths: (a) 3D model; (b) 2D model of flow paths.\cite{en15239180}}
    \label{model49}
\end{figure}

It involves an indirect liquid cooling system for a 2-parallel-8-series rectangular battery module. As shown in \Cref{model49}, the core component is a liquid cold plate with a multi-pass serpentine channel design. The system's operational strategy was optimized by selecting the flow rate ($\dot{m}$), inlet temperature ($T_{in}$), and average heat generation ($Q$) as the decision variables, with the objectives of minimizing the battery's average temperature ($T_{ave}$), maximum temperature difference ($\Delta T_{max}$), and the system's pressure drop($\Delta P$).

\textbf{Minimize:}
\begin{align}
f_1 = &\ 0.12313 - 0.1399x_1 + 0.9503x_2 + 0.9936x_3 \nonumber \\
& - 94.9058x_1 \cdot x_2 + 0.1523x_1 \cdot x_3 + 1.6680 \times 10^{-6}x_2 \cdot x_3 \nonumber \\
& - 177.3068x_1^2 + 4.8611 \times 10^{-7}x_2^2 - 0.0026x_1 \cdot x_2 \cdot x_3 \nonumber \\
& + 2379.7945x_1^2 \cdot x_2 + 3.4104x_1^2 \cdot x_3
\end{align}
\begin{align}
f_2 = &\ 0.0451 + 46.8433x_1 + 0.9258x_2 - 0.0036x_3 \nonumber \\
& - 88.6916x_1 \cdot x_2 - 1.7198x_1 \cdot x_3 + 0.0007x_2 \cdot x_3 \nonumber \\
& - 1947.0156x_1^2 + 1.1638 \times 10^{-6}x_2^2 + 0.000013x_3^2 \nonumber \\
& - 0.0271x_1 \cdot x_2 \cdot x_3 + 2190.9304x_1^2 \cdot x_2 \nonumber \\
& + 74.0804x_1^2 \cdot x_3 - 0.00006x_1 \cdot x_2^2
\end{align}
\begin{align}
f_3  = &\ 17.6338 + 10927.0303x_1 - 0.4822x_3 - 116.4686x_1 \cdot x_3 + 1865850x_1^2
\end{align}

\textbf{Subject to:}
\[
0.0249 \leq \dot{m} \leq 0.0498
\]
\[
5 \leq Q \leq 65
\]
\[
20 \leq T_{\text{in}} \leq 30
\]

\textbf{Where:}
\begin{align*}
x_1 &= \dot{m} \quad \text{(Flow rate, $kg\cdot s^{-1}$)} \\
x_2 &= Q \quad \text{(Inlet temperature, $W$)} \\
x_3 &= T_{in} \quad \text{(Average heat generation, $^\circ\text{C})$} \\
f_1 &= T_{ave}\quad \text{(Average temperature, $^\circ\text{C})$ )}\\
f_2 &= \Delta T_{max} \quad \text{(Maximum temperature difference, $^\circ\text{C})$)}\\
f_3 &= \Delta P \quad \text{(The system's pressure drop, Pa)} \\
\end{align*}

\subsection{U-Shaped Micro Heat Pipe Array Hybrid Cooling Model \protect\cite{ZENG2022119171}}
\begin{figure}[htbp]
    \centering
    \includegraphics[width=0.7\linewidth]{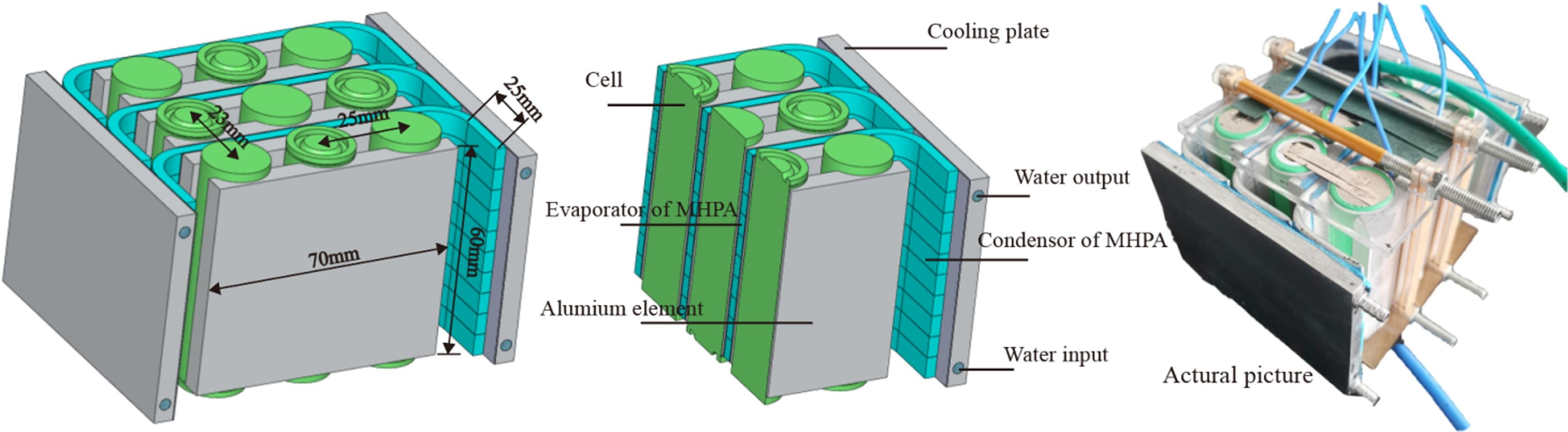}
    \caption{The hybrid battery module based on MHPA and liquid cooling.\cite{ZENG2022119171}}
    \label{model50}
\end{figure}

As shown in \Cref{model50}, it presents the optimized design of a thermal management structure for a cylindrical battery module, which integrates U-shaped micro heat pipe arrays (MHPAs) with liquid cold plates.

\textbf{Minimize:}
\begin{align}
f_1 = T_{max} =  &\ 81.01683 - 0.866333x_1 - 0.480056x_2 - 3.91737x_3 - 198.02083x_4 \nonumber \\
&+ 0.002183x_1x_2 + 0.009344x_1x_3 + 1.21563x_1x_4 + 0.0179x_2x_3 \nonumber \\
&- 0.391667x_2x_4 - 4.98750x_3x_4 + 0.011359x_1^2 \nonumber \\
&+ 0.002449x_2^2 + 0.225781x_3^2 + 1499.0625x_4^2
\end{align}

\begin{align}
f_2 = \Delta T =  &\ 27.31163 - 0.838542x_1 - 0.291767x_2 - 2.11342x_3 + 90.08333x_4 \nonumber \\
& + 0.002429x_1x_2 + 0.016469x_1x_3 + 0.140625x_1x_4 + 0.007333x_2x_3 \nonumber \\
&- 0.183333x_2x_4 - 1.94375x_3x_4 + 0.011346x_1^2 + 0.001407x_2^2 \nonumber \\ 
&+ 0.108792x_3^2 - 701.45833x_4^2
\end{align}

\begin{align}
f_3 = M =  &\ 0.569838 - 0.006269x_1 - 0.001207x_2 - 0.019008x_3 - 0.001125x_4 \nonumber \\
& + 0.000154x_1x_2 + 0.00145x_1x_3 + 0.000031x_1x_4 + 0.000451x_2x_3 \nonumber \\
&+ 0.000033x_2x_4 - 0.00025x_3x_4 + 7.70833\times 10^{-6}x_1^2 + 2.37037\times 10^{-7}x_2^2 \nonumber \\
&- 0.001966x_3^2 - 0.004167x_4^2
\end{align}

\textbf{Subject to:}
\[
20 \leq x_1 \leq 28
\]
\[
30 \leq x_2 \leq 60
\]
\[
2 \leq x_3 \leq 6
\]
\[
0.02 \leq x_4 \leq 0.06
\]

\textbf{Where:}
\begin{align*}
x_1 &= X_x \quad \text{(The battery spacing in X directions, mm)} \\
x_2 &= Hm \quad \text{(The height of the MHPA, mm)} \\
x_3 &= Wm \quad \text{(The aluminum conduction element thickness, mm)} \\
x_4 &= v \quad \text{(The coolant flow velocity, m/s)} \\
f_1 &= T_{max} \quad \text{(Maximum temperature of the battery, $^\circ{C}$)} \\
f_2 &= \Delta T \quad \text{(The temperature difference, $^\circ{C}$)} \\
f_3 &= M \quad \text{(The weight of the whole system, $kg$)} 
\end{align*}

\subsection{Dual-Bionic Cold Plate Structure Model \protect\cite{AN2024111541}}
\begin{figure}[htbp]
    \centering
    \includegraphics[width=0.7\linewidth]{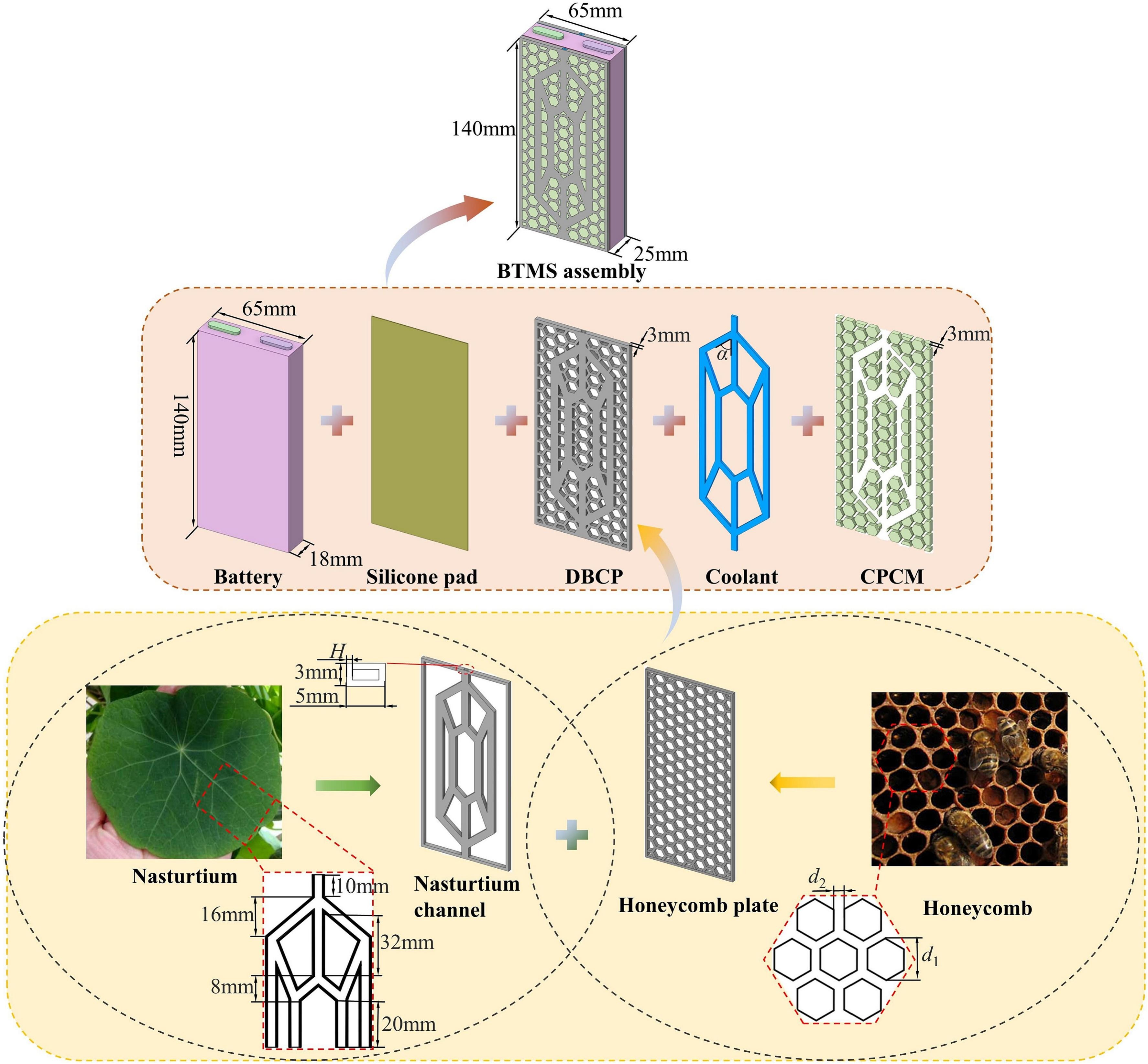}
    \caption{Dual bionic cold plate structure inspired by nasturtium veins and honeycomb patterns.\cite{AN2024111541}}
    \label{model51}
\end{figure}

It proposes a hybrid battery thermal management system with dual bionic cold plates for prismatic lithium batteries. As shown in \Cref{model51}, the system incorporates nasturtium vein-inspired channels and honeycomb structures.

\textbf{Minimize:}
\begin{align}
f_1 = &\ 46.61549 - 0.008793x_1 + 0.032083x_2 + 0.425x_3 - 3.08296x_4 \nonumber \\
& + 0.063125x_1x_3 - 0.06875x_2x_3 - 1.96875x_3x_4 - 0.000272x_1^2 \nonumber \\
& + 2.28962x_4^2
\end{align}
\begin{align}
f_2 = &\ 0.879167 + 0.013625x_1 + 0.16333x_2 + 5.53333x_3 + 1.48542x_4 \nonumber \\
& - 0.013125x_1x_4 - 0.045x_2x_4 - 3.125x_3^2 + 0.46875x_4^2
\end{align}
\begin{align}
f_3 = &\ 729.38807 + 0.3385x_2 - 1845.5756x_3 - 460.48423x_4 \nonumber \\
& + 724.5625x_3x_4 + 1124.54464x_3^2 + 48.57366x_4^2
\end{align}
\begin{align}
f_4 = &100-(\ 65.20346 + 0.179833x_1 + 0.719974x_2 - 25.15833x_3 \nonumber \\
& - 0.0022x_1x_2 + 0.1275x_1x_3 + 0.285x_2x_3 - 0.008856x_2^2)
\end{align}

\textbf{Subject to:}
\[
30 \leq x_1 \leq 70
\]
\[
40 \leq x_2 \leq 60
\]
\[
0.6 \leq x_3 \leq 1.0
\]
\[
0.2 \leq x_4 \leq 1.0
\]

\textbf{Where:}
\begin{align*}
x_1 &= \varepsilon \quad \text{(Porosity, \%)} \\
x_2 &= \alpha \quad \text{(Channel angle, $^\circ$)} \\
x_3 &= H \quad \text{(Channel wall thickness, mm)} \\
x_4 &= F_{in} \quad \text{(Inlet mass flow rate, g/s)} \\
f_1 &= T_{max} \quad \text{(Maximum temperature of the battery, $^\circ{C}$)} \\
f_2 &= \Delta T \quad \text{(The temperature difference, $^\circ{C}$)} \\
f_3 &= \Delta P \quad \text{(Inlet-outlet pressure drop, $Pa$)} \\
f_4 &= 100-\eta \quad \text{($\eta$ : Mass grouping rate, \%)}
\end{align*}

\subsection{Mini-Channel Cold Plate Optimization Model \protect\cite{KALKAN2022123949}}
\begin{figure}[htbp]
    \centering
    \includegraphics[width=0.7\linewidth]{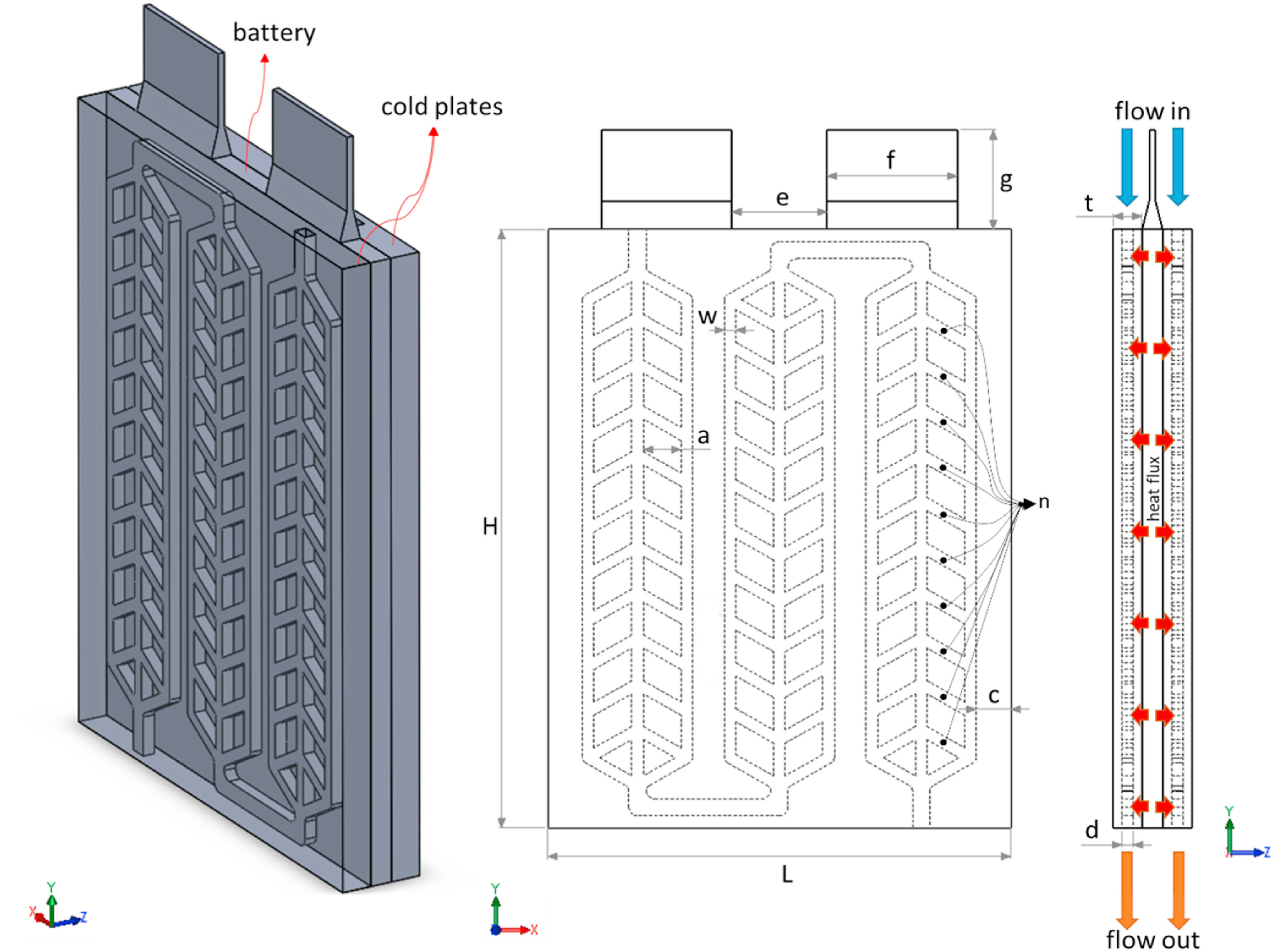}
    \caption{3D model and parameters of mini-channeled cold plates and LFP battery.\cite{KALKAN2022123949}}
    \label{model52}
\end{figure}

It involves a multi-objective optimization of a mini-channeled cold plate (MCCP) for Li-ion battery thermal management. As shown in \Cref{model52}, the system employs a sophisticated mini-channel design with optimized geometric parameters to enhance cooling performance while minimizing pressure drop.

\textbf{Minimize:}
\begin{align}
f_1 = &\ 1279 + 86x_1 - 443x_3 + 18x_4 + 9223x_5 + 21.98x_3^2 + 3887x_5^2 \nonumber \\
& - 314x_1x_5 + 92.8x_3x_5 - 566x_4x_5
\end{align}
\begin{align}
f_2 = &\ 52.096 - 0.463x_1 - 0.1498x_2 - 0.1197x_3 - 0.699x_4 - 4.432x_5 \nonumber \\
& + 1.481x_5^2 + 0.1150x_1x_4 + 0.01050x_2x_3 + 0.1033x_2x_5 - 0.0835x_3x_5
\end{align}
\begin{align}
f_3 = &\ 14.156 - 0.2739x_1 + 0.1121x_2 - 0.2305x_3 - 0.715x_4 + 0.091x_5 \nonumber \\
& - 0.00823x_2^2 + 0.007437x_3^2 + 0.0559x_4^2 - 0.8563x_5^2 + 0.06875x_1x_4 \nonumber \\
& + 0.007500x_2x_3 + 0.08333x_2x_5 - 0.06650x_3x_5 - 0.0425x_4x_5
\end{align}

\textbf{Subject to:}
\[
3 \leq x_1 \leq 5
\]
\[
10 \leq x_2 \leq 16
\]
\[
5 \leq x_3 \leq 15
\]
\[
3 \leq x_4 \leq 5
\]
\[
0.1 \leq x_5 \leq 1.1
\]

\textbf{Where:}
\begin{align*}
x_1 &= w \quad \text{(Channel width, mm)} \\
x_2 &= a \quad \text{(Distance between branches, mm)} \\
x_3 &= n \quad \text{(Number of crossovers in branches)} \\
x_4 &= d \quad \text{(Channel depth, mm)} \\
x_5 &= Q \quad \text{(Coolant flow rate, l/min)} \\
f_1 &= \Delta P \quad \text{(Pressure drop in channels, Pa)} \\
f_2 &= MBT \quad \text{(Maximum battery temperature, $^\circ{C}$)} \\
f_3 &= MTD \quad \text{(Maximum temperature difference on battery surface, $^\circ{C}$)}
\end{align*}

\subsection{Hybrid Air-PCM-Fin Thermal Management Model \protect\cite{KOSARI2024114392}}
\begin{figure}[htbp]
    \centering
    \includegraphics[width=0.7\linewidth]{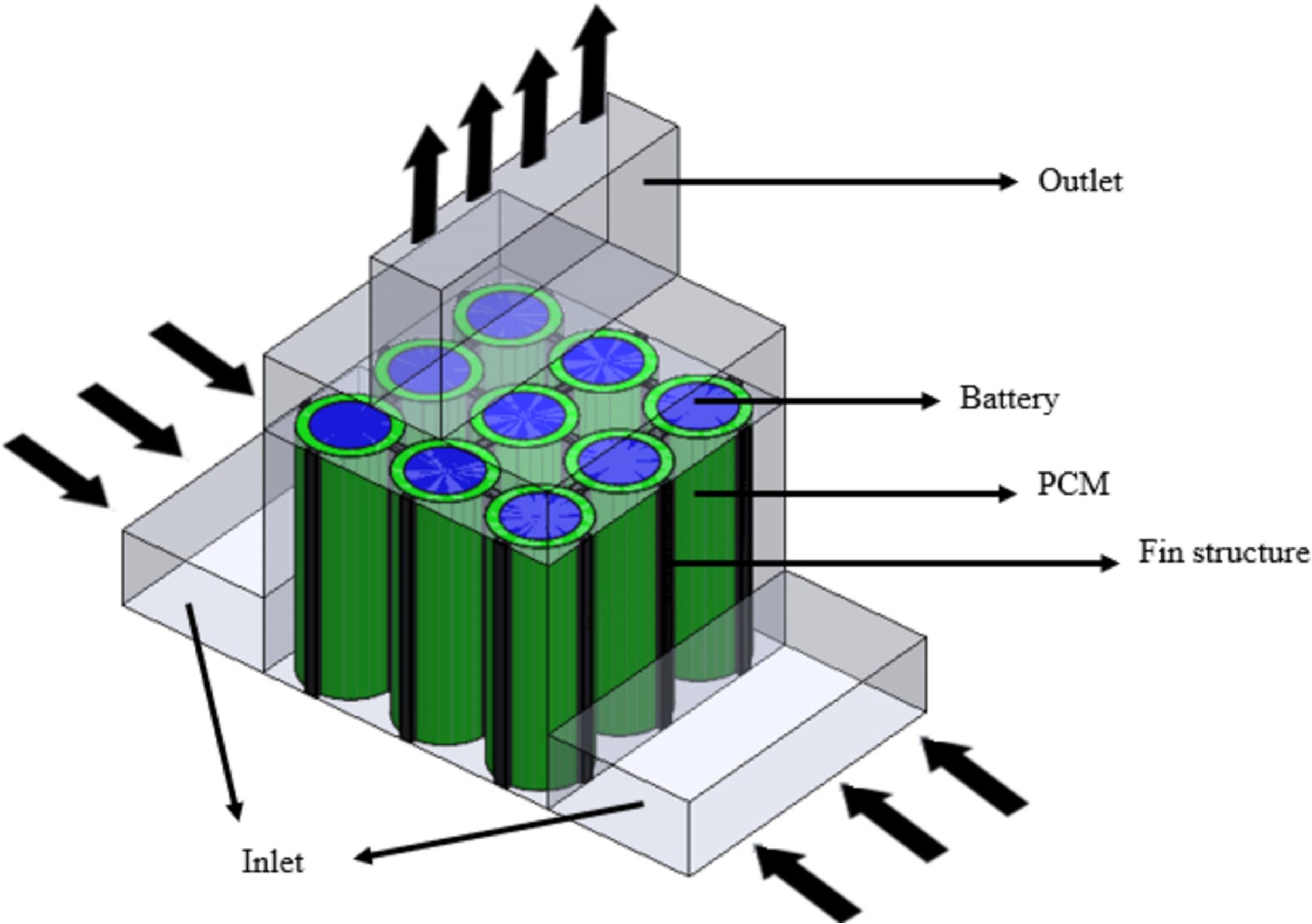}
    \caption{Schematic view of the hybrid BTMS design integrating forced air cooling, PCM with aluminum foam, and fin structures.\cite{KOSARI2024114392}}
    \label{model53}
\end{figure}

It involves a novel hybrid battery thermal management system (BTMS) combining forced air convection, aluminum fin structures, and phase change material (PCM) embedded with aluminum foam. As shown in \Cref{model53}, the system employs a Z-type airflow configuration for nine cylindrical 18650 lithium-ion batteries, with PCM enclosures enhanced by aluminum foam and fin structures to improve thermal conductivity and heat transfer efficiency.

\textbf{Minimize:}
\begin{align}
f_1 = &\ 4613.36904 + 209.50266x_1 - 3.50839x_2 - 31.98168x_3 \nonumber \\
& + 0.361111x_1x_2 - 0.768056x_1x_3 + 0.006746x_2x_3 \nonumber \\
& + 1.99043x_1^2 + 0.017178x_2^2 + 0.059304x_3^2
\end{align}

\begin{align}
f_2 = &\ -1833.39075 - 52.36404x_1 - 6.27092x_2 + 13.31269x_3 \nonumber \\
& - 0.253968x_1x_2 + 0.141667x_1x_3 + 0.024603x_2x_3 \nonumber \\
& + 1.51903x_1^2 - 0.009875x_2^2 - 0.02381x_3^2
\end{align}

\textbf{Subject to:}
\[
3 \leq x_1 \leq 4
\]
\[
0.5 \leq x_2 \leq 4
\]
\[
290.15 \leq x_3 \leq 300.15
\]

\textbf{Where:}
\begin{align*}
x_1 &= d \quad \text{(PCM thickness, mm)} \\
x_2 &= V_{inlet} \quad \text{(Inlet air velocity, m/s)} \\
x_3 &= T_{inlet} \quad \text{(Inlet air temperature, K)} \\
f_1 &= T_{max} \quad \text{(Maximum battery temperature, K)} \\
f_2 &= \Delta T_{max} \quad \text{(Maximum temperature difference between cells, K)} 
\end{align*}

\subsection{Tesla Valve with Spoiler Cooling Model \protect\cite{PENG2025124974}}
\begin{figure}[htbp]
    \centering
    \includegraphics[width=0.7\linewidth]{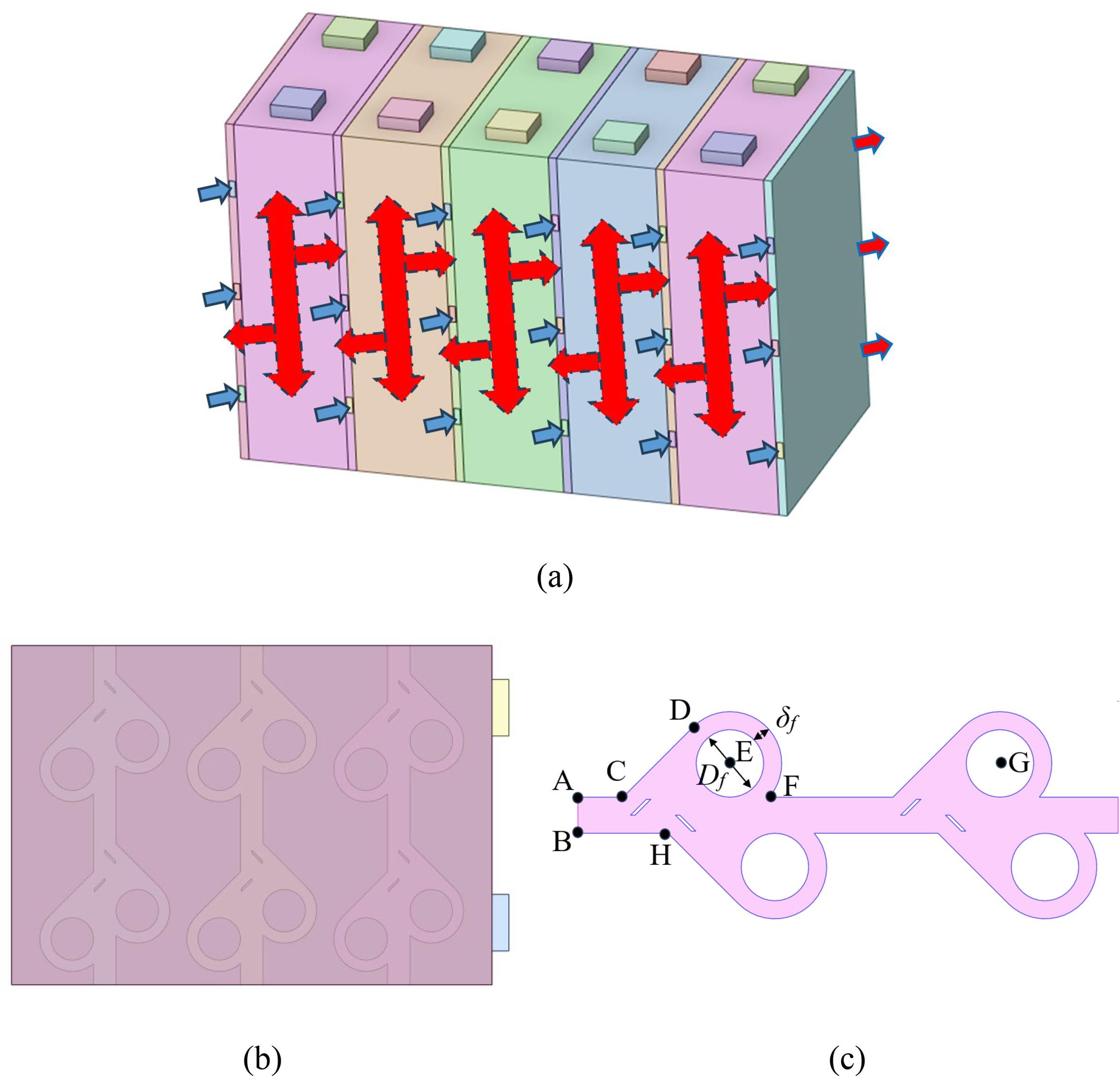}
    \caption{Liquid cooling system with Tesla flow channel. Three-dimensional view (a) of the cold plate model of a battery pack with five cells, arrows indicating the flow of coolant, the internal channel (b) of the cold plate containing the spoiler arrangement, and defining the channel parameters (c).\cite{PENG2025124974}}
    \label{model54}
\end{figure}

It involves an optimized liquid-cooled battery thermal management system featuring Tesla valve flow channels with integrated spoilers. As shown in \Cref{model54}, the system employs a novel design combining Tesla valve structures with strategically placed spoilers to enhance heat transfer performance through improved flow disturbance and turbulence generation, particularly suitable for high-capacity prismatic lithium-ion batteries.

\textbf{Minimize:}
\begin{align}
f_1 = &\ 1.90626712 \times 10^{-4} x_1^2 - 1.28889451 \times 10^{-3} x_1 x_2 \nonumber \\
& - 1.15404771 \times 10^{-1} x_1 x_3 + 3.13023994 \times 10^{-4} x_2^2 \nonumber \\
& + 8.56036434 \times 10^{-2} x_2 x_3 + 1.22088957 \times 10^{3} x_3^2 \nonumber \\
& + 8.59696045 \times 10^{-3} x_1 - 3.94856494 \times 10^{-4} x_2 \nonumber \\
& - 1.155495 \times 10^{2} x_3 + 32.2404955
\end{align}

\begin{align}
f_2 = &\ -7.90143347 \times 10^{-3} x_1^2 + 2.35570426 \times 10^{-2} x_1 x_2 \nonumber \\
& + 6.37633777 \times 10^{-2} x_1 x_3 + 4.92003546 \times 10^{-3} x_2^2 \nonumber \\
& - 5.68830689 \times 10^{-1} x_2 x_3 - 1.28949816 \times 10^{4} x_3^2 \nonumber \\
& - 3.19586641 \times 10^{-1} x_1 - 3.38066679 \times 10^{-1} x_2 \nonumber \\
& + 1.30319958 \times 10^{3} x_3 + 60.5388827
\end{align}

\textbf{Subject to:}
\[
5 \leq x_1 \leq 10
\]
\[
11.66 \leq x_2 \leq 16
\]
\[
0.01 \leq x_3 \leq 0.05
\]

\textbf{Where:}
\begin{align*}
x_1 &= L \quad \text{(Spoiler length, mm)} \\
x_2 &= X \quad \text{(Spoiler mounting position, mm)} \\
x_3 &= v \quad \text{(Coolant inlet flow velocity, m/s)} \\
f_1 &= T_{max} \quad \text{(Maximum battery temperature, $^\circ{C}$)} \\
f_2 &= Nu \quad \text{(Nusselt number)} 
\end{align*}

\subsection{Hybrid Manifold Channel Liquid Cooling Model \protect\cite{SUI2024123766}}
\begin{figure}[htbp]
    \centering
    \includegraphics[width=0.7\linewidth]{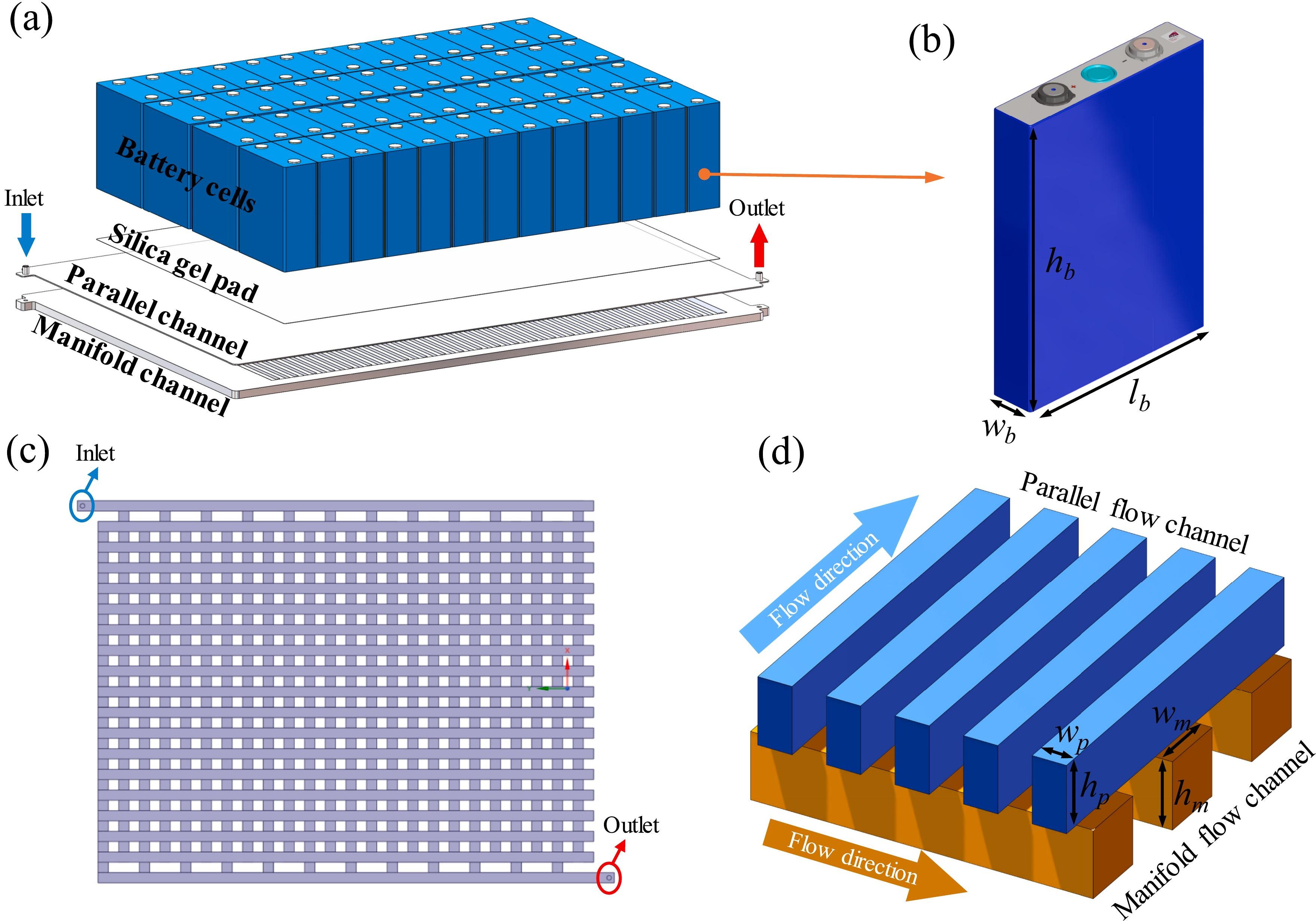}
    \caption{Schematic of the liquid cooling-based BTMS. (a) Diagram of lithium-ion battery pack; (b) Battery cell; (c) Flow domain of liquid cooling plate; (d) Design parameters of liquid cooling plate.\cite{SUI2024123766}}
    \label{model55}
\end{figure}

It involves a liquid cooling-based BTMS with hybrid manifold channels for a high-capacity 280 Ah LiFePO$_4$ battery pack. As shown in \Cref{model55}, the system features a cold plate incorporating both parallel and manifold channels positioned at the bottom of the battery pack.

\textbf{Minimize:}
\begin{align}
f_1 = &\ 72.10229 + 167.072x_1 - 54.5517x_2 - 596.305x_3 + 1387.328x_4 \nonumber \\
& + 1.45317x_1x_2 - 12.3493x_1x_3 - 20.6411x_1x_4 + 7.50918x_2x_3 \nonumber \\
& - 44.9616x_2x_4 + 98.18287x_3x_4 - 3.84157x_1^2 + 0.460898x_2^2 \nonumber \\
& + 75.53877x_3^2 + 1631.449x_4^2
\end{align}

\begin{align}
f_2 = &\ 24.85276 - 0.40694x_1 - 0.0449x_2 - 0.35387x_3 - 5.48901x_4 \nonumber \\
& + 0.000356x_1x_2 + 0.005188x_1x_3 + 0.002925x_1x_4 + 0.003644x_2x_3 \nonumber \\
& + 0.014112x_2x_4 - 0.00963x_3x_4 + 0.009482x_1^2 + 0.000145x_2^2 \nonumber \\
& + 0.006044x_3^2 + 0.956175x_4^2
\end{align}

\textbf{Subject to:}
\[
15 \leq x_1 \leq 25
\]
\[
20 \leq x_2 \leq 40
\]
\[
3 \leq x_3 \leq 5
\]
\[
1 \leq x_4 \leq 2
\]

\textbf{Where:}
\begin{align*}
x_1 &= w_p \quad \text{(Width of parallel channel, mm)} \\
x_2 &= w_m \quad \text{(Width of manifold channel, mm)} \\
x_3 &= h_p \quad \text{(Height of parallel channel, mm)} \\
x_4 &= v \quad \text{(Inlet velocity, m/s)} \\
f_1 &= \Delta P \quad \text{(Coolant pressure drop, Pa)} \\
f_2 &= \Delta T_{cells} \quad \text{(Temperature difference of battery modules, $^\circ\text{C}$)}
\end{align*}

\subsection{Bionic Lotus Leaf Channel Cooling Model \protect\cite{DONG2025134226}}
\begin{figure}[htbp]
    \centering
    \includegraphics[width=0.7\linewidth]{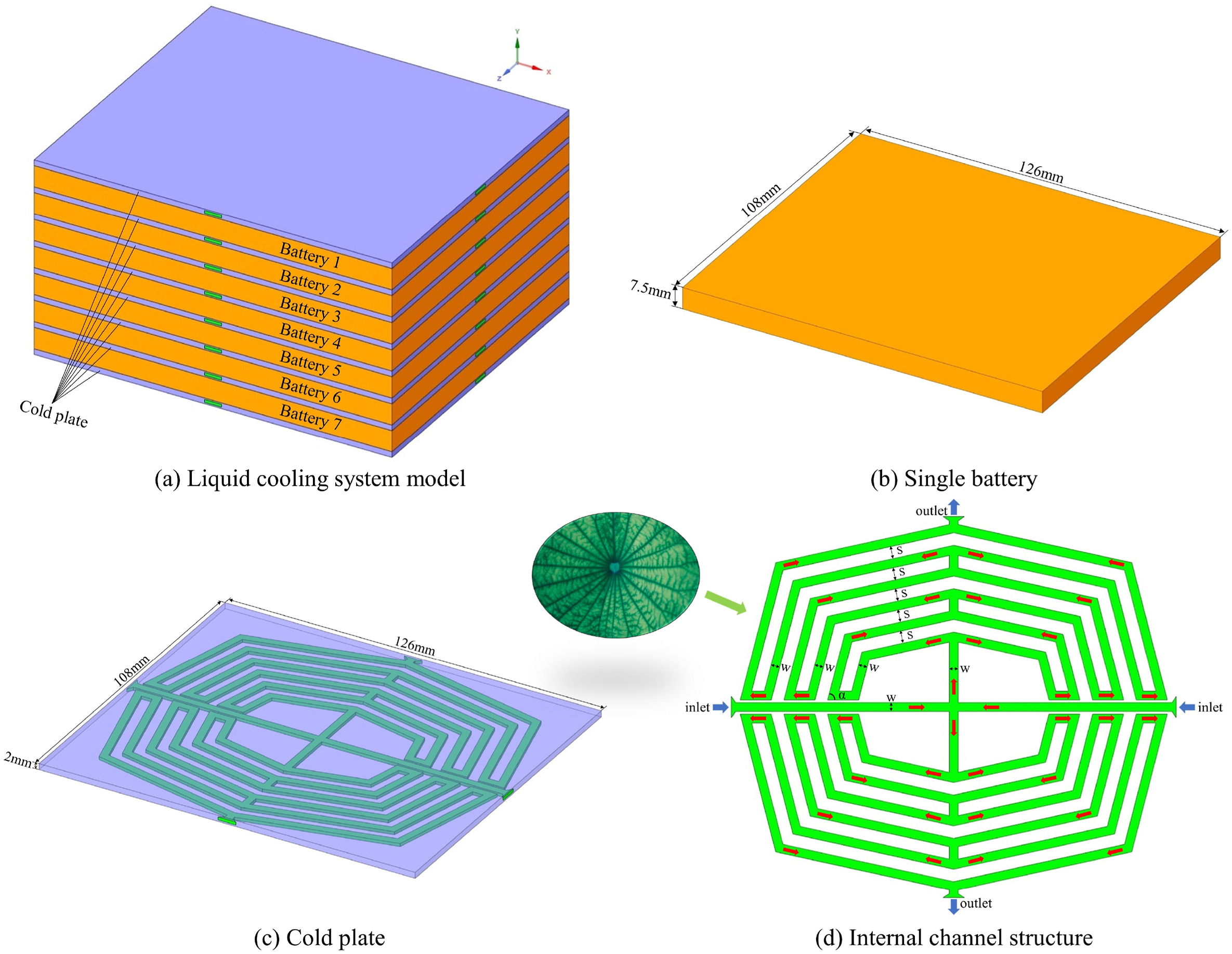}
    \caption{Schematic of the BTMS with novel bionic lotus leaf channels: (a) Liquid cooling system model; (b) Single battery; (c) Cold plate; (d) Internal channel structure.\cite{DONG2025134226}}
    \label{model56}
\end{figure}

This study proposes a novel bionic lotus leaf (NBLL) channel design for lithium-ion battery thermal management systems. As shown in \Cref{model56}, the system features innovative cooling plates inspired by lotus leaf vein patterns, with collecting and dispersing channels, two inlets, and two outlets. 

\textbf{Minimize:}
\begin{align}
f_1 = \Delta T_{max} = &\ 3.51 - 1.536x_1 + 0.062x_2 + 0.174x_3 - 0.0180x_4 \nonumber \\
& + 0.3677x_1^2 - 0.01724x_2^2 - 0.0278x_3^2 + 0.000069x_4^2 \nonumber \\
& + 0.00597x_1x_2 + 0.0049x_1x_3 - 0.00250x_1x_4 \nonumber \\
& + 0.00365x_2x_3 + 0.000716x_2x_4 - 0.00087x_3x_4
\end{align}

\begin{align}
f_2 = h = &\ 4940 - 4048x_1 - 70x_2 - 1367x_3 - 14x_4 + 760x_1^2 \nonumber \\
& + 76x_2^2 + 267x_3^2 - 0.11x_4^2 - 71x_1x_2 + 303x_1x_3 \nonumber \\
& + 46.8x_1x_4 - 65.0x_2x_3 - 3.91x_2x_4 + 0.3x_3x_4
\end{align}

\begin{align}
f_3 = \Delta P = &\ 27083 + 5034x_1 - 817x_2 - 6319x_3 - 460x_4 \nonumber \\
& + 923x_1^2 + 69x_2^2 + 2017x_3^2 + 3.00x_4^2 - 48x_1x_2 \nonumber \\
& - 2805x_1x_3 + 26.4x_1x_4 + 31x_2x_3 + 1.6x_2x_4 \nonumber \\
& - 21.3x_3x_4
\end{align}

\textbf{Subject to:}
\[
g_1 \leq 30.895
\]
\[
0.8 \leq x_1 \leq 2.0
\]
\[
3.0 \leq x_2 \leq 5.0
\]
\[
1.5 \leq x_3 \leq 3.0
\]
\[
75 \leq x_4 \leq 90
\]

where,
\begin{align*}
g_1 = &\ 47.3 - 9.970x_1 + 0.322x_2 + 0.545x_3 - 0.113x_4 \nonumber \\
& + 2.343x_1^2 - 0.0772x_2^2 - 0.106x_3^2 + 0.00042x_4^2 \nonumber \\
& + 0.0514x_1x_2 + 0.0932x_1x_3 - 0.00697x_1x_4 \nonumber \\
& + 0.0136x_2x_3 + 0.00219x_2x_4 - 0.00327x_3x_4
\end{align*}

\textbf{Where:}
\begin{align*}
x_1 &= M \quad \text{(Mass flow rate, g/s)} \\
x_2 &= S \quad \text{(Channel spacing, mm)} \\
x_3 &= W \quad \text{(Channel width, mm)} \\
x_4 &= \alpha \quad \text{(Channel angle, $^\circ$)} \\
f_1 &= \Delta T_{max} \quad \text{(Maximum temperature difference, $^\circ\text{C}$)} \\
f_2 &= h \quad \text{(Heat transfer coefficient, W/m$^2\cdot^\circ$C)} \\
f_3 &= \Delta P \quad \text{(Pressure drop, Pa)} \\
g_1 &= T_{max} \quad \text{(Maximum temperature constraint, $^\circ\text{C}$)} 
\end{align*}

\subsection{Topology-Optimized Liquid Cooling Plate Model \protect\cite{LIN2025119440}}
\begin{figure}[htbp]
    \centering
    \includegraphics[width=0.7\linewidth]{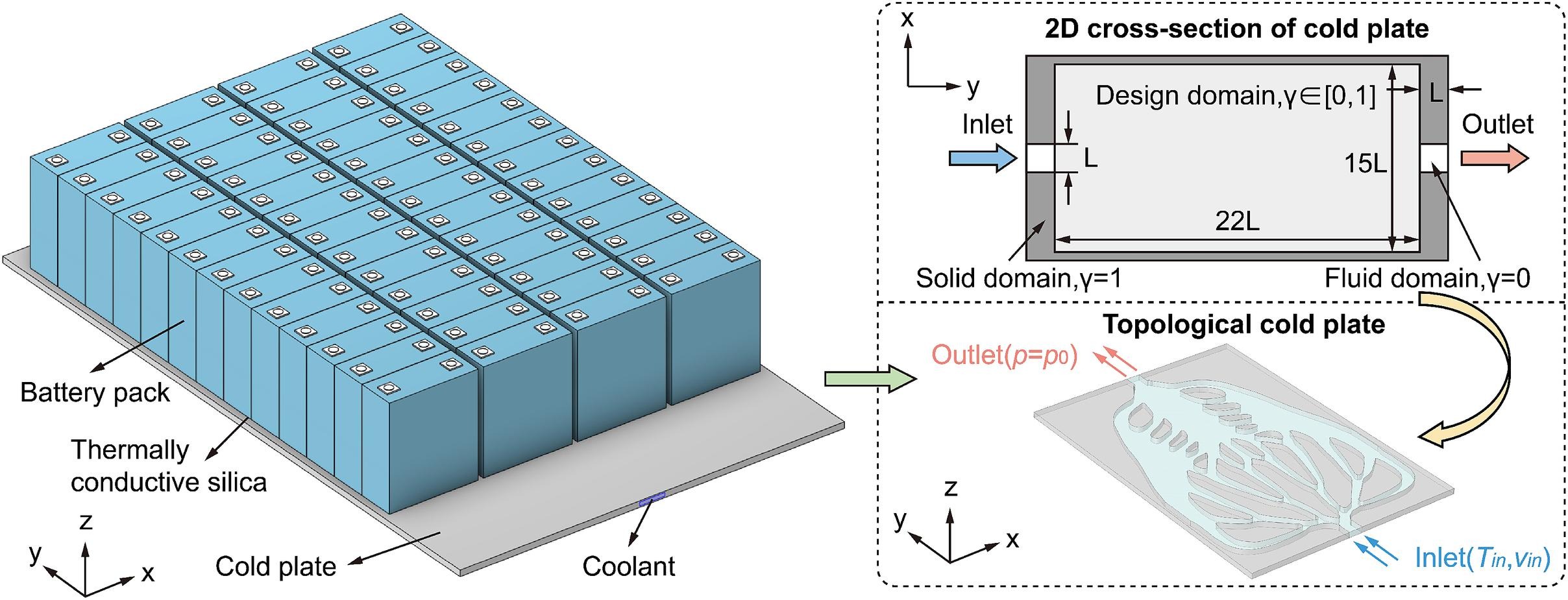}
    \caption{Schematic of the topology-optimized liquid cooling system: (a) 3D energy storage battery pack; (b) 2D topology optimization domain; (c) Structural evolution during optimization process.\cite{LIN2025119440}}
    \label{model57}
\end{figure}

As shown in \Cref{model57}, it involves a multi-objective topology optimization design for liquid-based cooling plates in 280 Ah prismatic energy storage batteries. The system employs a topology optimization approach to freely evolve the distribution of fluid domains within the cooling plate, considering the collaborative effects between heat dissipation, thermal uniformity, and flow resistance.

\textbf{Minimize:}
\begin{align}
f_1 = &\ 1\times10^4-(-33.674 + 16.72x_1 + 334.241x_2 + 178.136x_3 + 7.145x_1x_2 \nonumber \\
& -2.523x_1x_3 + 409.709x_2x_3 - 0.007x_1^2 - 565.378x_2^2 - 395.916x_3^2)
\end{align}
\begin{align}
f_2 = &\ 0.532 + 0.032x_1 - 1.786x_2 - 3.846x_3 - 0.008x_1x_2 \nonumber \\
& -0.024x_1x_3 + 2.365x_2x_3 - 4.06\times10^{-5}x_1^2 + 1.849x_2^2 + 3.524x_3^2
\end{align}
\begin{align}
f_3 = &\ 1\times10^{-8}(-68.397 + 0.784x_1 + 107.989x_2 + 1037.04x_3 \nonumber \\
& -1.435x_1x_2 - 6.57x_1x_3 - 1762.27x_2x_3 - 0.001x_1^2 \nonumber \\
& -115.21x_2^2 - 2301.1x_3^2 + 3.8013x_1x_2x_4 + 0.003x_1^2x_2 \nonumber \\
& +0.019x_1^2x_3 + 2.681x_1x_2^2 + 6.498x_1x_3^2 + 831.428x_2^2x_3 \nonumber \\
& +2481.38x_2x_3^2 + 1304.18x_3^3)
\end{align}

\textbf{Subject to:}
\[
\begin{cases}
50 < x_1 < 150 \\
0 < x_2 < 1 \\
0 < x_3 < 1 \\
x_2 + x_3 < 1
\end{cases}
\]

\textbf{Where:}
\begin{align*}
x_1 &= Re \quad \text{(Reynolds number)} \\
x_2 &= w_h \quad \text{(Heat dissipation weighting coefficient)} \\
x_3 &= w_u \quad \text{(Thermal uniformity weighting coefficient)} \\
f_1 &= 1e4 - Q_h \quad \text{($Q_h$: Heat dissipation rate, W/m)} \\
f_2 &= T_{uni} \quad \text{(Thermal uniformity, K}^2\text{)} \\
f_3 &=  Q_f \quad \text{(Flow energy dissipation, W/m)}
\end{align*}    

\subsection{Lightweight Hybrid PCM-Liquid Cooling Model \protect\cite{WU2020120495}}
\begin{figure}[htbp]
    \centering
    \includegraphics[width=0.7\linewidth]{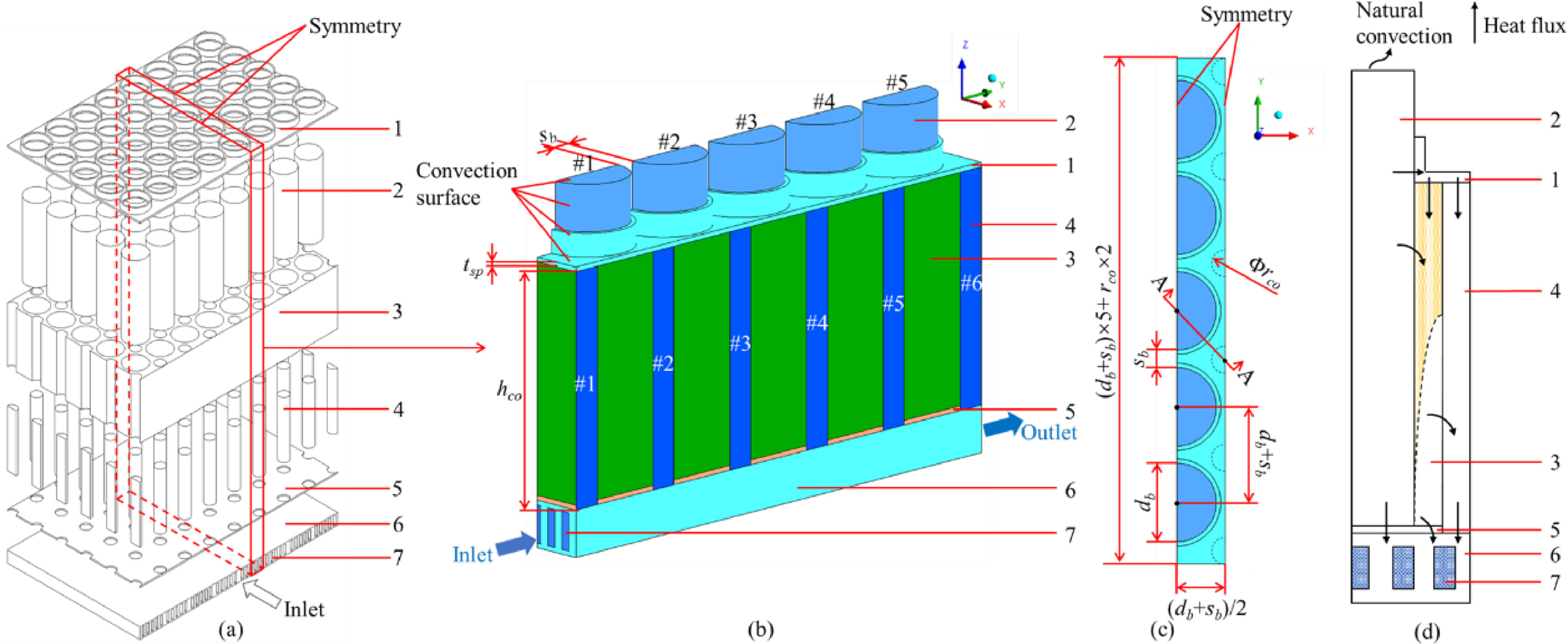}
    \caption{Hybrid cooling configuration for battery module: (a) Overall system schematic; (b) Numerical model of battery module; (c) Representative battery submodule; (d) Heat flow paths in A-A cross-section.\cite{WU2020120495}}
    \label{model58}
\end{figure}

This study presents a structural optimization of a light-weight battery module based on hybrid liquid cooling with high latent heat PCM. As shown in \Cref{model58}, the system features a novel cooling structure consisting of a minichannel cold plate at the bottom, thermal columns connecting to lateral sides through a heat spreading plate, and high latent heat PCM filled among batteries.

\textbf{Minimize:}
\begin{align}
f_1 = &\ 71.3037 - 3.9144x_1 - 1.6696x_2 - 0.6683x_3 - 0.4216x_4 \nonumber \\
                & - 0.0499x_1x_2 + 0.0706x_1x_3 + 0.0405x_1x_4 - 0.0255x_2x_3 \nonumber \\
                & + 0.0162x_2x_4 + 0.0017x_3x_4 + 0.0888x_1^2 + 0.3451x_2^2 \nonumber \\
                & + 0.0337x_3^2 + 0.0018x_4^2
\end{align}

\begin{align}
f_2 = &\ 39.3227 - 2.8721x_1 - 0.7085x_2 - 0.5715x_3 - 1.1052x_4 \nonumber \\
                 & + 0.0289x_1x_2 + 0.0185x_1x_3 + 0.0271x_1x_4 + 0.0106x_2x_4 \nonumber \\
                 & + 0.2723x_1^2 + 0.0535x_3^2 + 0.0095x_4^2
\end{align}

\begin{align}
f_3 = &\ 49.0164 - 10.2271x_1 + 0.6715x_2 + 1.6666x_3 - 0.0715x_4 \nonumber \\
               & + 0.1196x_1x_2 + 0.1486x_1x_3 + 0.2467x_1x_4 + 0.6145x_2x_3 \nonumber \\
               & + 0.1989x_3x_4 + 1.7310x_1^2 + 0.3715x_3^2
\end{align}

\textbf{Subject to:}
\begin{align}
35 \leq & f_1 \leq 50 \\
0 \leq & f_2 \leq 5 \\
0 \leq & f_3 \leq 150 \\
1 \leq & x_1 \leq 5 \\
0.2 \leq & x_2 \leq 2 \\
2 \leq & x_3 \leq 6 \\
45 \leq & x_4 \leq 55
\end{align}

\textbf{Where:}
\begin{align*}
x_1 &= r_{co} \quad \text{(Thermal column radius, mm)} \\
x_2 &= t_{sp} \quad \text{(Heat spreading plate thickness, mm)} \\
x_3 &= s_b \quad \text{(Battery spacing, mm)} \\
x_4 &= h_{co} \quad \text{(Thermal column height, mm)} \\
f_1 &= T_{max} \quad \text{(Maximum temperature, $^\circ\text{C}$)} \\
f_2 &= \Delta T \quad \text{(Maximum temperature difference, $^\circ\text{C}$)} \\
f_3 &= m_{ts} \quad \text{(Thermal system mass, g)}
\end{align*}

\subsection{MHPA-PCM-Liquid Triple Hybrid Cooling Model \protect\cite{XIE2023121341}}
\begin{figure}[htbp]
    \centering
    \includegraphics[width=0.7\linewidth]{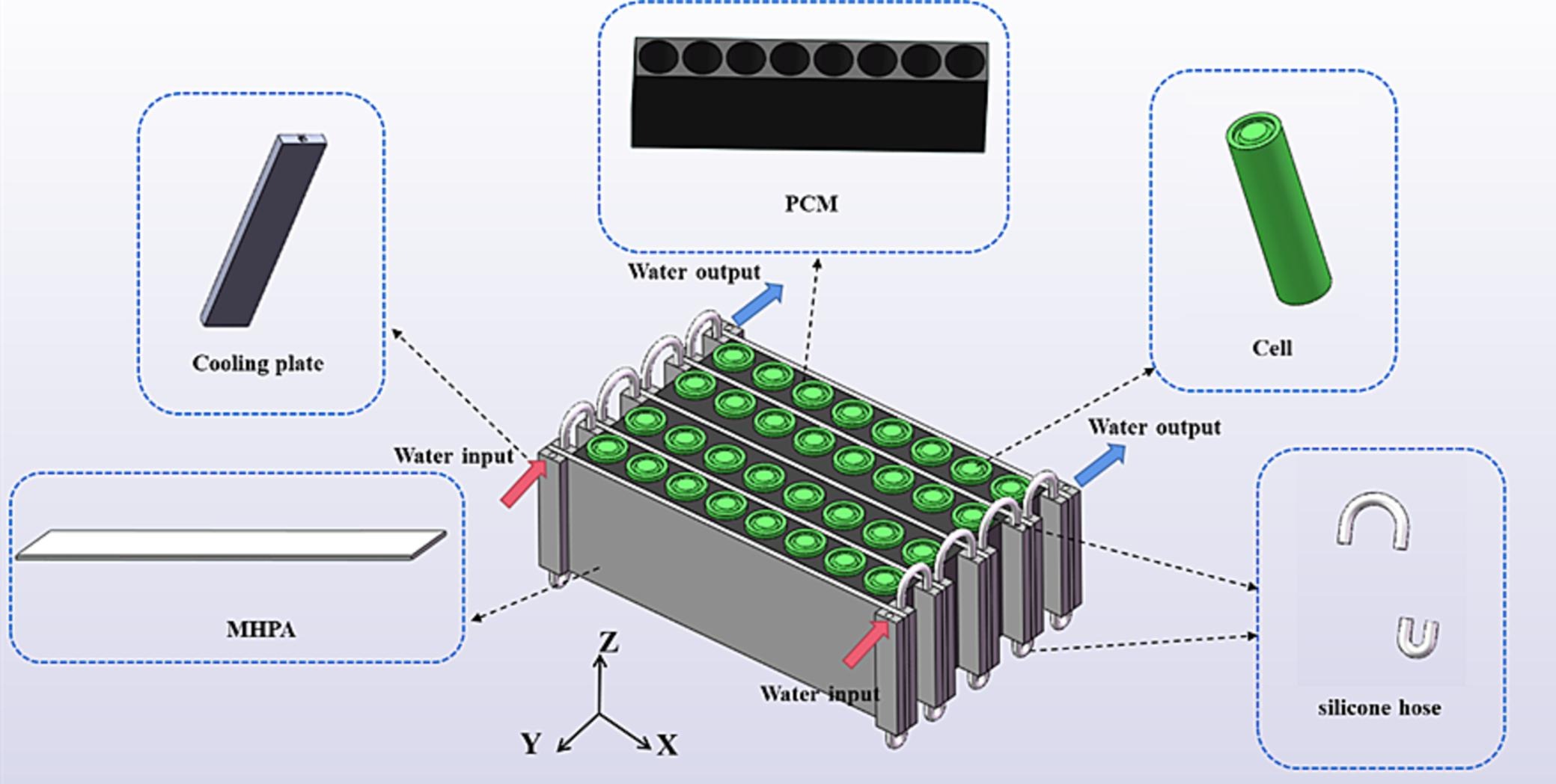}
    \caption{Schematic of the designed hybrid BTMS with PCM, cooling plates, and MHPA.\cite{XIE2023121341}}
    \label{model59}
\end{figure}

It involves a hybrid thermal management system based on MHPA/PCM/liquid cooling for lithium-ion batteries. As shown in \Cref{model59}, the system features a novel design combining micro heat pipe array (MHPA), phase change material (PCM), and liquid cooling plates for efficient thermal management of 4 × 8 cylindrical battery modules.

\textbf{Minimize:}
\begin{align}
f_1 = &\ -2.43064 + 0.243045x_1 + 0.677324x_2 + 0.737592x_3 - 0.18472x_4 \nonumber \\
                & - 0.02606x_1x_2 + 0.022396x_1x_3 + 0.004529x_1x_4 - 0.02294x_2x_3 \nonumber \\
                & + 0.00055x_2x_4 + 0.004813x_3x_4 - 0.11394x_1^2 - 0.00341x_2^2 \nonumber \\
                & - 0.01697x_3^2 + 0.000302x_4^2
\end{align}

\begin{align}
f_2 = &\ 96.58241 + 0.1578x_1 - 1.25045x_2 - 2.3343x_3 - 0.12563x_4 \nonumber \\
                & - 0.07581x_1x_2 + 0.049944x_1x_3 - 0.00313x_1x_4 + 0.027531x_2x_3 \nonumber \\
                & - 0.00033x_2x_4 + 0.005656x_3x_4 + 0.028913x_1^2 + 0.011411x_2^2 \nonumber \\
                & + 0.017132x_3^2 - 0.00048x_4^2
\end{align}

\textbf{Maximize:}
\begin{align}
f_3 = &\ 300-( 277.272 - 0.51318x_1 - 2.63203x_2 - 2.6598x_3 - 0.50029x_4 \nonumber \\
           & - 0.00716x_1x_2 + 0.056038x_1x_3 - 0.0782x_1x_4 - 0.01772x_2x_3 \nonumber \\
           & + 0.001278x_2x_4 + 0.008648x_3x_4 + 0.046377x_1^2 + 0.024511x_2^2 \nonumber \\
           & + 0.017802x_3^2 + 0.001049x_4^2)
\end{align}

\textbf{Subject to:}
\begin{align}
2 \leq & x_1 \leq 4 \\
20 \leq & x_2 \leq 28 \\
23 \leq & x_3 \leq 29 \\
30 \leq & x_4 \leq 60
\end{align}

\textbf{Where:}
\begin{align*}
x_1 &= W_m \quad \text{(MHPA thickness, mm)} \\
x_2 &= X_x \quad \text{(Battery spacing in X direction, mm)} \\
x_3 &= Y_y \quad \text{(Battery spacing in Y direction, mm)} \\
x_4 &= H_m \quad \text{(MHPA height, mm)} \\
f_1 &= \Delta T \quad \text{(Temperature difference, $^\circ\text{C}$)} \\
f_2 &= T_{max} \quad \text{(Maximum temperature, $^\circ\text{C}$)} \\
f_3 &= 300-ED \quad \text{(ED: Energy density, Wh/kg)}
\end{align*}

\subsection{Passive Flow Regulation \& Uneven Channel Model \protect\cite{SHI2025135331}}
\begin{figure}[htbp]
    \centering
    \includegraphics[width=0.7\linewidth]{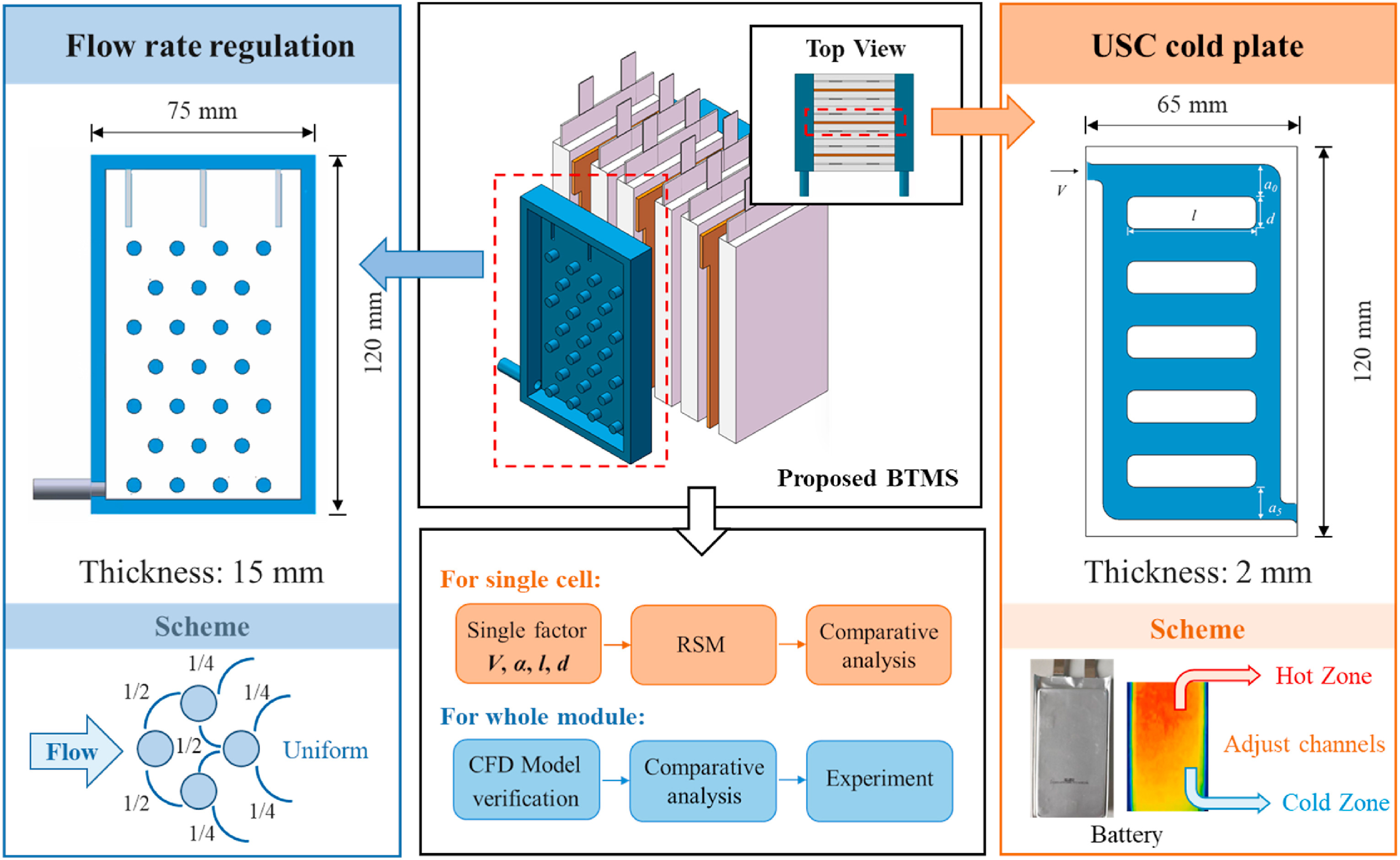}
    \caption{Design and dimensions of the proposed BTMS.\cite{SHI2025135331}}
    \label{model60}
\end{figure}

It involves a BTMS based on passive flow rate regulation and unequally spaced channels. As shown in \Cref{model60}, the system features an optimized cold plate with Unequally Spaced Channel(USC) design and a passive flow regulation structure inspired by the Galton board for uniform coolant distribution across battery modules.

\textbf{Minimize:}
\begin{align}
f_1= &\ 59.83219 + 0.12937x_1 - 0.98175x_2 - 13.24975x_3 - 7.77460x_4 \nonumber \\
                & - 0.00427x_1x_2 + 0.28775x_1x_3 - 0.06762x_1x_4 - 0.06180x_2x_3 \nonumber \\
                & - 0.00740x_2x_4 + 3.93312x_3x_4 + 0.00630x_1^2 + 0.013343x_2^2 \nonumber \\
                & + 12.80125x_3^2 + 2.46094x_4^2
\end{align}

\begin{align}
f_2 = &\ 20.98648 + 0.18112x_1 - 0.55020x_2 - 2.53808x_3 - 6.25006x_4 \nonumber \\
                 & - 0.00439x_1x_2 + 0.10915x_1x_3 - 0.027650x_1x_4 - 0.19525x_2x_3 \nonumber \\
                 & + 0.026362x_2x_4 + 1.37000x_3x_4 + 0.00245x_1^2 + 0.00796x_2^2 \nonumber \\
                 & + 1.83333x_3^2 + 1.60591x_4^2
\end{align}

\textbf{Subject to:}
\begin{align}
25 \leq & f_1 \leq 40 \\
0 \leq & f_2 \leq 8.5 \\
0.8 \leq & x_4 \leq 1.6 \\
0 \leq & x_3 \leq 0.3 \\
35 \leq & x_2 \leq 50 \\
0 \leq & x_1 \leq 20
\end{align}

\textbf{Where:}
\begin{align*}
x_1 &= d \quad \text{(Separator width, mm)} \\
x_2 &= l \quad \text{(Separator length, mm)} \\
x_3 &= \Delta\alpha \quad \text{(Common difference coefficient of channel spacing)} \\
x_4 &= V \quad \text{(Coolant flow rate, L/min)} \\
f_1 &= T_{max} \quad \text{(Maximum battery temperature, $^\circ\text{C}$)} \\
f_2 &= \Delta T \quad \text{(Temperature difference, $^\circ\text{C}$)}
\end{align*}
\section{Conclusion}
This paper has presented a comprehensive benchmark suite specifically designed for multi-objective optimization in battery thermal management system (BTMS) design.  Recognizing the limitations of Synthetic Benchmark Problems (SBPs) that often contain unrealistic properties, we have developed a collection of 12 real-world constrained multi-objective optimization problems (RWCMOPs) derived from recent BTMS research.  These problems cover a diverse range of cooling technologies including serpentine channel liquid cooling, micro heat pipe arrays, bionic structures, mini-channel cold plates, hybrid PCM systems, and topology-optimized designs.

The key contribution of this work lies in providing readily implementable mathematical formulations based on accurate surrogate models that efficiently capture complex thermal-fluid interactions.  Each problem is presented with clearly defined decision variables, objective functions, and constraint boundaries, making this benchmark suite immediately applicable for evaluating constrained multi-objective evolutionary algorithms.  The problems encompass essential trade-offs in BTMS design, including thermal performance (maximum temperature and temperature uniformity), hydraulic performance (pressure drop), and structural characteristics (system weight and size).

This benchmark suite addresses a significant gap in the current landscape of optimization test problems by offering realistic problems derived from practical engineering applications, computationally efficient surrogate models replacing expensive CFD simulations, diverse problem characteristics with varying difficulty levels, standardized mathematical formulations for easy implementation, and comprehensive coverage of multiple BTMS technologies and design approaches.

For future work, several directions are envisioned to extend the utility and impact of this benchmark suite.  First, comprehensive experimental evaluation using state-of-the-art constrained multi-objective optimization algorithms will be conducted to establish baseline performance results.  Second, the development of a standardized ranking scheme will facilitate fair and consistent comparison of different algorithms.  Third, the benchmark suite may be expanded to include more problems from emerging BTMS technologies and different battery configurations.  Finally, the integration of these problems into popular optimization platforms and frameworks will enhance their accessibility to the research community.

This benchmark suite provides a valuable resource for researchers working on evolutionary algorithms and optimization methods, offering a more realistic and relevant testing environment compared to traditional synthetic problems.  We believe it will contribute to advancing the state-of-the-art in both optimization algorithms and battery thermal management design by enabling more robust and meaningful performance assessments.
\bibliographystyle{unsrt}
\bibliography{ref}
\end{document}